\documentclass{article}
\usepackage{ijcai26}

\usepackage{times}
\usepackage{soul}
\usepackage{url}
\usepackage[hidelinks]{hyperref}
\usepackage[utf8]{inputenc}
\usepackage[small]{caption}
\usepackage{graphicx}
\usepackage{amsmath}
\usepackage{amsthm}
\usepackage{booktabs}
\usepackage{algorithm}
\usepackage{algorithmic}
\usepackage[switch]{lineno}

\title{
The Sword, Shield, and Achilles' Heel: Characterizing the Linguistic Inductive Bias of Large Language Models for Spatial Reasoning in Navigation Planning
}

\author{
    Xudong Zhang$^1$
    \and
    Jian Yang$^2$\footnote{Corresponding author}\and
    Shengkai Wang$^3$\and
    Jiangpeng Tian$^2$\and 
    Shaowen Chen$^1$\and
    Xian Wei$^1$\and
    Ke Li$^2$\and
    Xiong You$^2$
    \affiliations
    $^1$East China Normal University, China\\
    $^2$Information Engineering University, China\\
    $^3$Zhengzhou University, China\\
    \emails
    71275900012@stu.ecnu.edu.cn,\\
    \{jian.yang, shengkai.wang, xian.wei\}@tum.de,\\
    \{tjpeng2011, 15952613852, like19771223, youarexiong\}@163.com
}

\begin{document}

\maketitle

\begin{abstract}
Large Language Model (LLM)-based navigation systems commonly construct explicit spatial representations (e.g., topological graphs, semantic raster maps) and translate them into textual descriptions as LLMs' inputs. However, the linguistic structures of such text-based spatial representations and the choices of contextual features (e.g., topology, geometry) they contain are often treated as neutral engineering decisions rather than key factors that shape LLMs' behavior. To fill the gap, we propose a dual-interventional framework that disentangles linguistic structures from different contextual cues to evaluate the linguistic inductive bias of LLMs for navigation planning. In the framework, representation intervention varies the linguistic format and the degree of linguistic compression, clarifying when linguistic representations support or inhibit navigation planning. Context intervention, combined with contextual feature combination and conflict probing, explicitly clarifies the preferences and weaknesses of LLMs when processing different contextual cues. Experiments across diverse spatial reasoning tasks and multiple model scales reveal a consistent pattern: topological information is a sturdy shield and the backbone of robust planning; linguistic format is a double-edged sword whose effect depends on model size, task demands, and the compression level; and semantic information is a fatal Achilles' heel---incorrect semantic cues can systematically derail the planning process. Overall, our study shows that effective text-based spatial representations in LLM-based navigation should preserve topological integrity, calibrate representational compression to model capacity, and ensure semantic correctness, rather than simply adopting a single representation. Our code is publicly available at \url{https://github.com/jonesdong150/LLM-Navigation-Inductive-Bias}.
\end{abstract}

\section{Introduction}

Spatial intelligence is a central capability of embodied artificial intelligence, particularly in navigation tasks where agents must plan goal-directed paths rather than rely on reactive obstacle avoidance \cite{Deitke2020,Savva2019Habitat}.
Recently, Large Language Models (LLMs) have been adopted as high-level navigation planners, responsible for reasoning over symbolic spatial representations and generating navigation sub-goals \cite{Wei2022,Yao2022ReAct,Wang2023Voyager}.
Most existing systems follow a hierarchical architecture, in which LLMs infer abstract navigation decisions from explicit spatial inputs—such as topological graphs or semantic maps—while lower-level controllers execute actions \cite{Shah2023LMNav,Huang2023VLMaps,Rana2023SayPlan}.

Despite rapid progress, a fundamental question remains insufficiently explored:
\emph{How does the linguistic representation of serialized spatial inputs shape LLMs' navigation reasoning behavior?}
Prior work largely treats serialization as a neutral engineering step, converting structured spatial representations into text to facilitate planning, with limited systematic analysis of how linguistic structure and contextual cues influence reasoning outcomes \cite{Chen2024MapGPT}.

From the perspective of LLMs, Transformer-based reasoning operates over token sequences by selectively attending to and integrating evidence across long contexts \cite{Vaswani2017}.
Different linguistic organizations—such as flat descriptions, hierarchical groupings, or clustered representations—induce distinct attention patterns and evidence salience.
Moreover, spatial cues differ in their \emph{sequential compatibility}: some cues, such as topological connectivity, align naturally with sequential processing, whereas others, such as coordinate-based geometric descriptions, do not.
When combined with representational choices, these differences can systematically bias navigation reasoning even under information-equivalent spatial conditions \cite{Liu2023LostInMiddle,Fatemi2023Talk2Graph}.

Cognitive map theory provides complementary methodological guidance.
Spatial cognition research emphasizes that navigation relies on multiple spatial information channels and that hierarchical and clustered organizations are fundamental properties of spatial knowledge \cite{Tolman1948,Montello1998Framework,Montello2001SpatialCognition,MontelloSas2006Wayfinding,HirtleJonides1985}.
Rather than attributing biological cognition to LLMs, we adopt these principles strictly as a design perspective to guide controlled interventions over linguistic structure and contextual cue accessibility.

\begin{figure*}[t]
  \centering
  \includegraphics[width=1.0\textwidth]{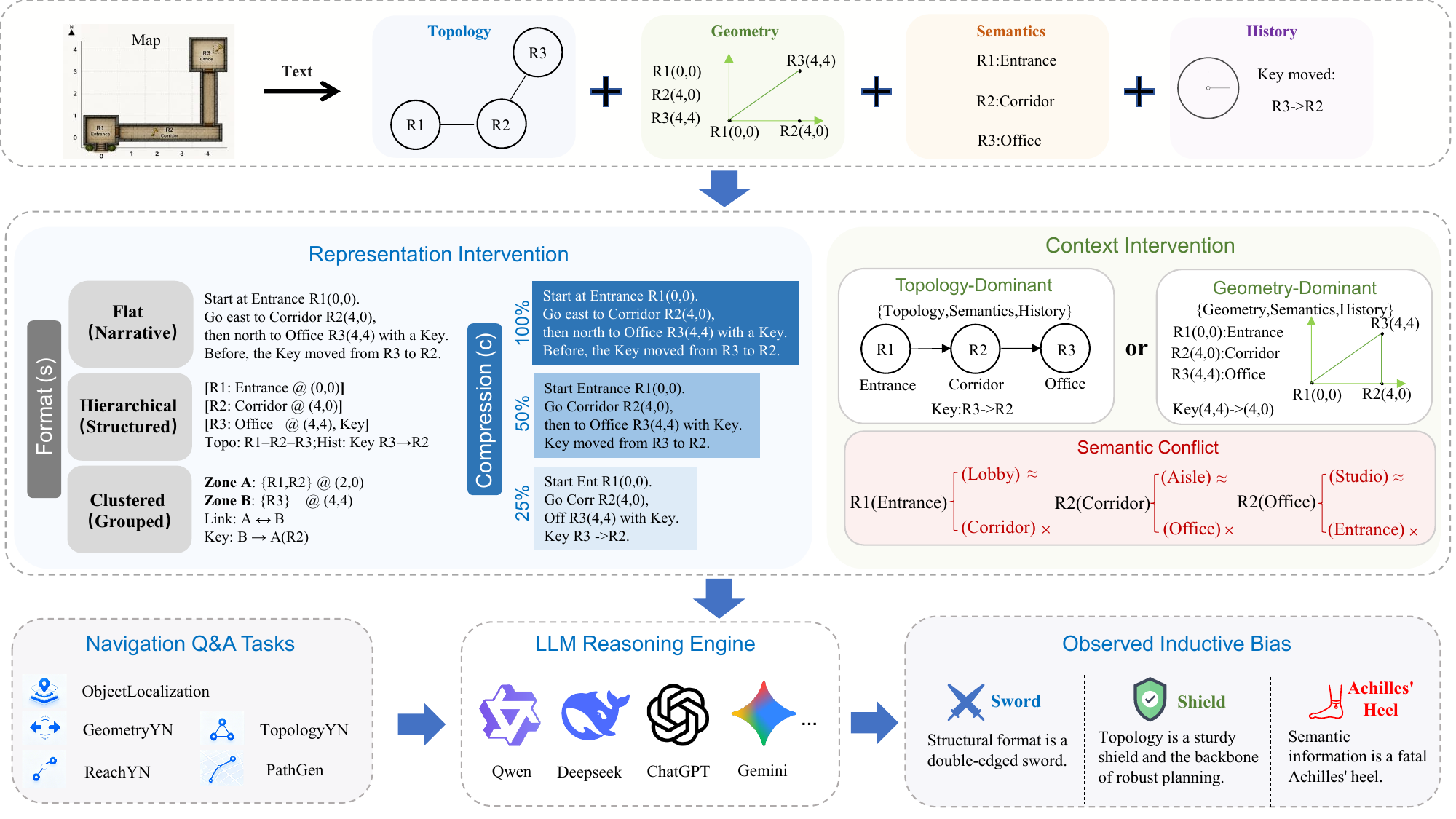}
  \vspace{-1mm}
  \caption{Unified framework overview.
Representation intervention manipulates linguistic organization and compression under information-equivalent settings through Flat, Hierarchical, and Clustered formats.
Context intervention controls the accessibility and consistency of Topology, Geometry, Semantics, and History cues through dominance and conflict probing.
The framework reveals three characteristic inductive-bias signatures in LLM-based navigation reasoning:
the Sword (structure-dependent representation effects),
the Shield (topological robustness),
and the Achilles' Heel (semantic vulnerability).}
  \label{fig:paradigm}
\end{figure*}

Guided by these perspectives, we propose a controlled experimental framework consisting of \emph{representation intervention} and \emph{context intervention} (Fig.~\ref{fig:paradigm}).
Unlike prompt engineering approaches that primarily target performance optimization \cite{Kojima2022,Zhou2023NavGPT}, our framework enforces \emph{information equivalence}: the underlying spatial reality is held fixed while linguistic format, compression level, contextual cue combinations, and cue conflicts are systematically manipulated.
We evaluate these factors across five spatial reasoning tasks, disentangling the effects of linguistic structure from those of contextual cues and exposing models' implicit prioritization strategies and vulnerabilities.

Across experiments and model scales, we observe a consistent pattern of linguistic inductive biases.
Topological information acts as a robust \emph{shield} supporting stable planning.
Linguistic format functions as a \emph{double-edged sword}, whose benefit depends on model capacity, task demands, and compression.
Semantic information emerges as an \emph{Achilles' heel}: incorrect semantic cues can systematically derail navigation reasoning.
Together, these findings highlight the importance of preserving topological integrity, calibrating representational compression, and ensuring semantic correctness in text-based spatial representations for LLM-based navigation.

\section{Related Work}

\subsection{Explicit Spatial Representations for LLM-Based Navigation}

LLMs are increasingly integrated into navigation as high-level planners, using topological graphs or semantic maps for environment serialization \cite{Huang2022InnerMonologue,Ahn2022,Chen2024MapGPT,Zhang2025MapNav,Zhou2023NavGPT}. Most prior methods adopt an engineering-oriented perspective, treating these descriptions as neutral data carriers. However, empirical findings remain mixed: linearizing complex structures—such as occupancy grids or hierarchical maps \cite{Thrun2005,Kuipers2000,Latombe1991,Liang2022CodeAsPolicies,Wang2023NLGraph}—into token sequences can introduce inductive biases that conflict with Transformer pre-training \cite{Fatemi2023Talk2Graph,Zhang2024GraphLLM}, potentially degrading performance for larger models in complex environments \cite{Valmeekam2023}.

Unlike these performance-centric studies, we treat linguistic organization and contextual cues as explicit experimental variables. This allows us to systematically analyze how specific representational choices shape LLM-based navigation reasoning rather than merely optimizing for system-level metrics.

\subsection{Cognitive Maps}

Cognitive science suggests that navigation relies on multiple spatial channels (topology, geometry, landmarks) organized through hierarchical and clustered structures \cite{Tolman1948,Montello2001SpatialCognition,HirtleJonides1985,Montello1998Framework}. While recent AI research invokes cognitive maps as a metaphor to probe LLMs' world models \cite{Gurnee2023SpaceTime,Bubeck2023}, most studies remain descriptive and fail to operationalize these principles for controlled analysis. Consequently, it remains unclear how specific spatial cues influence navigation reasoning.

In contrast, our work operationalizes key cognitive map principles into a dual-interventional framework. By translating these principles into manipulable variables—through representation intervention and context conflict probing—we enable a systematic analysis of how LLMs prioritize, integrate, and override spatial cues in language-mediated navigation.

\section{Methodology}

This section introduces a dual-intervention framework for examining reasoning behavior in LLM-based navigation planning.
Rather than optimizing system performance, our goal is to characterize how LLMs respond to controlled variations in spatial representations and contextual cues.
We treat \textit{linguistic format}, \textit{degree of linguistic compression}, \textit{contextual cue combination}, and \textit{contextual cue conflict} as externally controllable variables.
These variables are manipulated through \textit{representation and context interventions}, which constitute the primary interface through which serialized spatial information is supplied to the LLM reasoning engine (Figure~\ref{fig:paradigm}).

\subsection{Representation Intervention: Linguistic Format and Compression}
\label{sec:repr_intervention}

To analyze how the \textit{representational structure} of serialized language influences navigation reasoning, we design representation interventions along two fundamental dimensions:
\textit{linguistic format} ($s$) and \textit{degree of linguistic compression} ($c$).
This design choice is motivated by the observation that language-based representations differ primarily in (i) how spatial information is \emph{structurally organized}, and (ii) how much information is \emph{compressed} into a fixed-length token sequence.
Moreover, the choice of linguistic format is informed by principles from cognitive map theory, where hierarchical and clustered organizations are regarded as fundamental structures for organizing spatial knowledge.

Formally, let $E$ denote a spatial environment and $I(E)$ its ground-truth \emph{contextual spatial cues}.
A representation intervention is defined as a mapping
\begin{equation}
    \mathcal{F}_{s,r}: I(E) \rightarrow \mathcal{T}_{s,r},
\end{equation}
where $s \in \{\text{Flat, Hier, Clus}\}$ specifies the linguistic format (as shown in the upper middle part of Figure~\ref{fig:paradigm}) and
$r \in \{100\%, 50\%, 25\%\}$ denotes the degree of information retention rate (100\% indicates the original, uncompressed text). (as illustrated in the compression branch of Figure~\ref{fig:paradigm} for the Flat example).

As illustrated in Figure~\ref{fig:paradigm}, we enforce strict Information Equivalence:
for any combination of $(s,c)$, the resulting token sequence $\mathcal{T}_{s,c}$ encodes an identical set of \emph{contextual spatial cues}, spanning Topology,Geometry,Semantics and History.
Thus, representation intervention modifies only the \emph{structure} and \emph{textual density} of the language interface, while keeping the underlying spatial information invariant.

\paragraph{Linguistic Format Intervention (Regime R1).}
To isolate the independent effect of \emph{linguistic format}, we vary the structural organization of spatial information while holding spatial complexity and contextual spatial cues constant (as illustrated in the upper part of Figure~\ref{fig:paradigm}):
\begin{itemize}
    \item \textit{Flat}: A continuous natural language narrative that describes the environment in a linear prose form.
    This format aligns with dominant pre-training data distributions (e.g., web text and documents) of contemporary LLMs such as LLaMA~\cite{touvron2023llama,grattafiori2024llama3} and Qwen~\cite{bai2023qwen,qwen2024qwen2}.
    \item \textit{Hierarchical}: A structured representation that organizes spatial information into nested levels (e.g., regions, rooms, and objects) with explicit structural markers.
    \item \textit{Clustered}: A functional grouping strategy that partitions entities and locations into localized zones based on relational or functional proximity.
\end{itemize}
Crucially, all three formats contain exactly the same underlying spatial facts—namely, identical contextual spatial cues in Topology,Geometry,Semantics and History.
The only difference lies in the representational structure of the interface.
This controlled setup transforms serialized language into a measurable variable, enabling precise quantification of how structural organization affects LLM-based navigation planning.

\paragraph{Linguistic Compression Intervention (Regime R2).}
To examine how textual density interacts with representational structure, we manipulate the \emph{information retention rate} ($r \in \{100\%, 50\%, 25\%\}$) within each format:
\begin{itemize}
\item \textit{Full Retention (100\%)}: Fully articulated descriptions with complete syntactic scaffolding.
\item \textit{Moderate Retention (50\%)}: Semi-structured representations that remove redundant linguistic fillers while preserving all contextual spatial cues.
\item \textit{Low Retention (25\%)}: Highly condensed symbolic encodings that maximize information density within the token sequence.
\end{itemize}
Importantly, compression intervention preserves all contextual spatial cues and modulates only the \emph{textual density} of these cues within a fixed attention budget, thereby enabling controlled analysis of how compression interacts with model capacity during planning.

\subsection{Context Intervention: Cue Dominance and Semantic Conflict}
\label{sec:context_intervention}

Beyond representational form, we further decompose spatial information into a set of explicit contextual cue dimensions:
\[
\mathcal{C} = \{\text{Topology}, \text{Geometry}, \text{Semantics}, \text{History}\}.
\]
Here, Topology refers to node connectivity; Geometry involves coordinate sets; Semantics represents object/room labels; and History tracks the temporal sequence. As illustrated in Figure~\ref{fig:paradigm}, context intervention selectively manipulates the availability and consistency of these dimensions to probe the model's reasoning pillars.

\paragraph{Contextual Cue Combination (Regime C1 Results).}
Drawing from the duality of spatial representations in robotics, we construct two non-conflicting regimes to test \emph{cue sufficiency}. Both regimes provide sufficient information for navigation but rely on distinct reasoning backbones:
\begin{itemize}
    \item \textit{Topology-Dominant}: The model operates on an explicit connectivity graph (nodes and edges). This setup mirrors the \textit{Spatial Semantic Hierarchy} in cognitive robotics~\cite{Kuipers2000} and recent LLM-based topological planners~\cite{Zhou2023NavGPT,Shah2023LMNav}, where navigation is treated as a graph search problem.
    \item \textit{Geometry-Dominant}: The model operates on coordinate sets without explicit links. This mirrors metric Simultaneous Localization and Mapping (SLAM) systems~\cite{Thrun2005} or vector-based neural planners~\cite{Zhang2017NeuralSLAM,Parisotto2017NeuralMap}, requiring the LLM to infer adjacency and traversability implicitly via Euclidean distance calculations.
\end{itemize}

Comparing performance across these regimes reveals the model's intrinsic \emph{inductive bias}: does the LLM reason more effectively over explicit graph structures (Topo) or implicit coordinate spaces (Geom)?

\paragraph{Contextual Conflict Probing (Regime C2 Results).}
To expose system vulnerabilities, we introduce a targeted \emph{Semantic Conflict} probe (see the semantic-conflict branch in Figure~\ref{fig:paradigm}). This design simulates \textit{perceptual aliasing}, a critical failure mode in real-world embodied AI where vision systems fail to distinguish between visually similar environments~\cite{Cadena2016past}.

We inject this conflict specifically into the Topology-Dominant regime—typically considered the robust "shield" of navigation—to conduct an adversarial stress test on this most robust component, asking whether even its topological support harbors exploitable vulnerabilities.
\begin{itemize}
    \item \textit{Structural Truth (The Shield)}: The topological graph remains perfect, with unique identifiers (IDs) (e.g., $ID_{R2} \neq ID_{R3}$) and correct connectivity.
    \item \textit{Semantic Ambiguity (The Trap)}: We deliberately inject duplicate semantic labels for distinct nodes (e.g., labeling both the \textit{Guest Bedroom} and \textit{Master Bedroom} simply as "Bedroom").
\end{itemize}
This probe forces a competition between strict logic and semantic intuition. A robust reasoner should distinguish nodes by their unique IDs or topological context (History), whereas a fragile reasoner will be misled by the semantic ambiguity. We hypothesize that this semantic confusion constitutes the "Achilles' heel" of LLM spatial reasoning.

\section{Experiments}
To ensure that the observed behaviors reflect systematic
trends rather than stochastic noise, our dual-interventional ex-
periments (comprising R1, R2, C1, and C2) generated a total
of 57,500 queries. This extensive evaluation scale provides
high-fidelity insights into the pillars of navigation reasoning
in LLMs.

\subsection{Scene Datasets and Navigation Tasks}
\label{sec:exp_setup}

We evaluate our proposed interventions on three curated synthetic scene sets (Set-A, B, and C). While large-scale open-source navigation datasets such as Matterport3D or Habitat-based benchmarks offer visual diversity, they are unsuitable for our study because their annotations are typically \textit{context-incomplete} and do not allow for the strict \textit{variable isolation} required by our intervention axes~\cite{Cadena2016past,Chen2024MapGPT}. Specifically, existing datasets often lack synchronized and independent control over Topology, Geometry, and History, making it impossible to enforce the Information Equivalence necessary to isolate representational effects from information availability.

Our synthetic design addresses these gaps:
\begin{itemize}
    \item \textit{Set-A}: 10 low-complexity scenes for linguistic format intervention (R1).
    \item \textit{Set-B}: 10 scenes with complexity gradients ($G_1$--$G_5$), used for linguistic compression (R2).
    \item \textit{Set-C}: 10 scenes specifically designed for context interventions (C1: Dominance and C2: Conflict).
\end{itemize}

The experimental regimes and the detailed complexity schedule for Set-B are summarized in Table~\ref{tab:merged_setup}. Across all sets, we evaluate five tasks spanning core spatial cognitive requirements:
(1) ObjectLocation: Local entity grounding~\cite{Huang2023VLMaps};
(2) GeometryYN: Validating spatial relations (e.g., "left/right", "A is north of B") through spatial reasoning~\cite{Shah2023LMNav};
(3) TopologyYN: Verifying connectivity from linear text~\cite{Wang2023NLGraph};
(4) ReachYN: Multi-hop reachability analysis;
(5) PathGen: Integrated sequential planning~\cite{Ahn2022}.

\begin{table}[t]
\centering
\small
\setlength{\tabcolsep}{5pt}
\renewcommand{\arraystretch}{1.1}
\caption{\textbf{Experimental Overview.} (Top) Intervention regimes and dataset assignments. All regimes are evaluated across 5 tasks. (Bottom) Detailed complexity schedule for Set-B used in Regime R2.}
\label{tab:merged_setup}

\begin{tabular}{l c c l}
\hline
\textbf{Regime} & \textbf{Scene Set} & \textbf{Complexity} & \textbf{Variables} \\
\hline
R1 & Set-A & Fixed (Low) & Format ($s$) \\
R2 & Set-B & $G_1$--$G_5$ & $s \times c$ \\
C1 & Set-C & Fixed (Med) & Cue mix \\
C2 & Set-C & Fixed (Med) & Semantic conflict \\
\hline
\end{tabular}

\vspace{2pt}

\begin{tabular}{c c c c l}
\multicolumn{5}{c}{\textit{Set-B Complexity Details ($G_1$--$G_5$)}} \\
\hline
\textbf{Grad} & \textbf{\#Rms} & \textbf{\#Objs} & \textbf{Hist.} & \textbf{Cognitive Load} \\
\hline
$G_1$ & 6  & 4  & 2--3   & Basic cognition \\
$G_2$ & 8  & 6  & 4--5   & Basic cognition \\
$G_3$ & 10 & 8  & 6--7   & Structured reasoning \\
$G_4$ & 12 & 10 & 8--9   & Structured reasoning \\
$G_5$ & 14 & 12 & 10--11 & Extreme reasoning \\
\hline
\end{tabular}
\end{table}

\subsection{Experimental Models}
\label{sec:models}

Our model selection spans a range of reasoning capabilities and deployment constraints, covering both open and closed models. This tiered strategy is crucial for characterizing how linguistic inductive biases evolve with model capacity~\cite{Kaplan2020Scaling}, and is also motivated by practical deployment considerations, offering actionable guidance for embodied systems.

\begin{itemize}
    \item \textit{Deployment-Friendly (Small)}: Ultra-lightweight models ($\leq$4B) optimized for on-device execution, including Qwen3 (0.6B, 1.7B, 4B) and Llama-3.2 (1B, 3B).
    \item \textit{Deployment-Feasible (Medium)}: Models that typically require high-performance edge hardware or lightweight cloud instances, including Llama-3.1-8B, Qwen3-14B, and Qwen3-32B.
    \item \textit{High-Intelligence (Large)}: Application Programming Interface (API)-accessible frontier models as capability ceilings, including ChatGPT-5.2 and Gemini-2.5. All experiments use a temperature of 0.0 for deterministic reproducibility.
\end{itemize}

\subsection{Representation Intervention: Linguistic Format and Compression}

Following the representation intervention defined in \S\ref{sec:repr_intervention}, we evaluate how varying linguistic format and compression modulates navigation reasoning under controlled, information-equivalent conditions.

\begin{figure}[t]
    \centering
    \includegraphics[width=0.90\columnwidth]{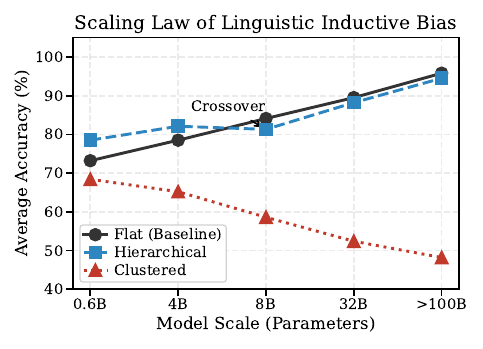}
    \caption{Scaling behavior of linguistic inductive bias (Set-A). A crossover occurs across model scales: small models favor hierarchical formats, while larger models prefer flat descriptions.}
    \label{fig:seta_structure}
\end{figure}

\paragraph{Structural Inductive Effects (R1 Results).}
On Set-A, the linguistic format intervention demonstrates that representational structure functions as a \textit{conditional inductive bias} that scales with model capacity (Figure~\ref{fig:seta_structure}). For deployment-friendly models ($\leq$4B), the \texttt{Hierarchical} format consistently outperforms the \texttt{Flat} baseline. From a model-mechanism perspective, explicit hierarchical markers pre-organize spatial facts into attention-aligned blocks, reducing the burden on limited-capacity Transformers to infer long-range relational structure and enabling more stable evidence aggregation during navigation planning.

As the parameter scale increases, this structural advantage diminishes and a distinct \textit{crossover} point emerges within the deployment-feasible model interval. At this stage, the \texttt{Flat} format begins to match or surpass the \texttt{Hierarchical} format. This suggests that larger models possess the emergent capacity to internally reconstruct spatial relations, rendering rigid structural scaffolds unnecessary or even counterproductive due to inductive interference~\cite{Fatemi2023Talk2Graph}. At the high-parameter ceiling (e.g., frontier LLMs), the preference for \texttt{Flat} descriptions becomes more pronounced.

These findings offer clear engineering guidance: for edge-side deployment where small models are preferred, hierarchical structuring provides essential "scaffolding" for navigation; whereas for cloud-based large models, standard flat descriptions are more effective and avoid structural constraints.

In contrast, the \texttt{Clustered} format remains consistently inferior across the entire parameter spectrum. This persistent degradation aligns with a \emph{fragmentation effect}~\cite{Liu2023LostInMiddle}: partitioning spatial information into loosely connected clusters impairs the model's ability to integrate evidence for navigation planning. This leads to a practical engineering insight: designers should avoid adopting clustered structures when representing spatial environments for LLM-based navigation planning. By holding spatial complexity and contextual spatial cues constant, R1 isolates linguistic structure as a key variable, revealing how structural biases interact with Transformer scale to shape navigation behavior.

\begin{figure*}[t]
  \centering
  \includegraphics[width=0.96\textwidth]{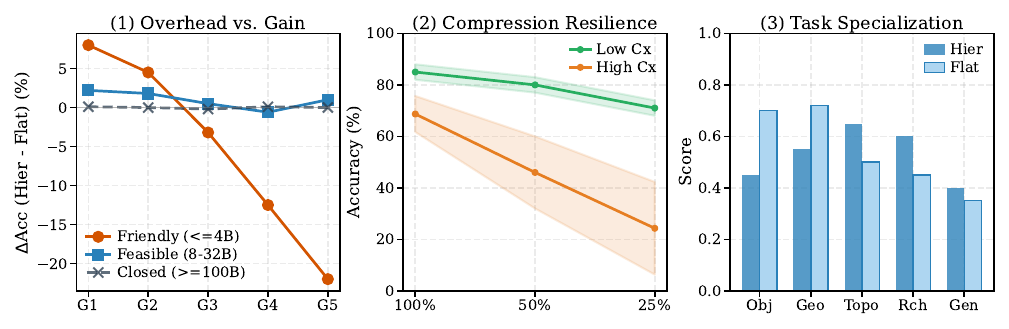}
  \caption{Complexity-dependent boundaries and task-specific utility (R2).
  (1) Inductive gain $\Delta Acc$ reveals a scaffolding-to-overhead reversal for small models.
  (2) Compression resilience shows a clear “threshold collapse” in high-complexity regimes.
  (3) Task-wise accuracy at $G_5$ highlights functional specialization between formats.}
  \label{fig:r2_combined}
\end{figure*}

\paragraph{Complexity-Dependent Structural Reversal (R2 Results).}
R2 complements R1 by introducing a spatial complexity gradient ($G_1$--$G_5$) and controlled linguistic compression to probe the operational boundaries of structural inductive biases (Figure~\ref{fig:r2_combined}). Note that based on R1 results, the \texttt{Clustered} format is excluded from R2's boundary analysis due to its consistent inferiority. Unless otherwise specified, results report mean accuracy aggregated across scenes and models within corresponding tiers.

\textit{(1) Scaffolding-to-Overhead Reversal:}
Figure~\ref{fig:r2_combined}(1) plots the relative gain $\Delta Acc = Acc_{\texttt{Hier}} - Acc_{\texttt{Flat}}$. For deployment-friendly models ($\leq$4B), $\Delta Acc$ is positive at low complexity ($G_1$--$G_2$) but reverses sharply beyond $G_3$, yielding a substantial negative margin at $G_5$. Mechanistically, this reflects a shift from \emph{structural help} to \emph{structural overhead}: at low spatial load, hierarchical markers align attention to local blocks; as complexity increases, nested structures expand token length and introduce formatting burden, competing with evidence aggregation under a fixed attention budget. In contrast, deployment-feasible and closed-source models ($\geq$4B) show near-zero mean $\Delta Acc$ across all levels, indicating they can recover relational structure from natural text more robustly.

\textit{Engineering Guidance:} For edge deployment with limited-capacity models, use hierarchical formatting only for simple environments; for complex spatial tasks, switch to flat descriptions to minimize structural overhead.

\textit{(2) Dual Impact of Linguistic Compression:}
Figure~\ref{fig:r2_combined}(2) reports compression resilience across complexity levels. At low complexity ($G_1$--$G_2$), performance degrades moderately as compression increases, showing high redundancy tolerance. In contrast, high-complexity regimes ($G_3$--$G_5$) exhibit a \emph{threshold collapse}, where aggressive compression (below 50\%) triggers a non-linear drop in accuracy. Mechanistically, aggressive compression removes error-correcting cues essential for attention-based retrieval in complex scenes.

\textit{Engineering Guidance:} While spatial descriptions can be heavily compressed for simple maps to save tokens, high-complexity environments require maintaining at least 50\%–100\% of the original text to preserve critical connectivity signals.

\textit{(3) Structural Selectivity across Navigation Tasks:}
Figure~\ref{fig:r2_combined}(3) decomposes task-wise performance at high complexity ($G_5$). For clarity in visualization, task names are abbreviated as follows: ObjLoc (ObjectLocation), TopoYN (TopologyYN), PathG (PathGen), ReachYN (ReachabilityYN), and GeomYN (GeometryYN).

Results show that \texttt{Hier} performs best on \textit{TopoYN} and \textit{ReachYN}, acting as a topological anchor that stabilizes multi-hop evidence aggregation. Conversely, \texttt{Flat} is strongest on \textit{ObjLoc} and \textit{GeometryYN}, where lower token overhead favors fine-grained attribute binding and local relation checks.

\textit{Engineering Guidance:} The choice of representation format should be task-specific: prioritize hierarchical structures for pure connectivity or reachability tasks, and use flat descriptions for attribute-intensive or local geometric checks.

\subsection{Context Intervention: Cue Dominance and Semantic Conflict}
\label{sec:results_context_intervention}

Unless otherwise specified, all metrics in this section are averaged across all models and scenes under the corresponding intervention conditions, aiming to highlight systemic trends while suppressing confounding variance. Based on the framework defined in Sec.~\ref{sec:context_intervention}, we analyze the model's reasoning pillars through two regimes.

\paragraph{Cue Dominance (Regime C1 Results).}
In this regime, we evaluate the model's preference between topological connectivity and geometric coordinates. We define the Dominance Score ($S_{dom}$) to quantify this bias:
\begin{equation}
    S_{dom} = Acc_{topo} - Acc_{geom}
\end{equation}

\begin{figure}[t]
  \centering
  \includegraphics[width=0.90\columnwidth]{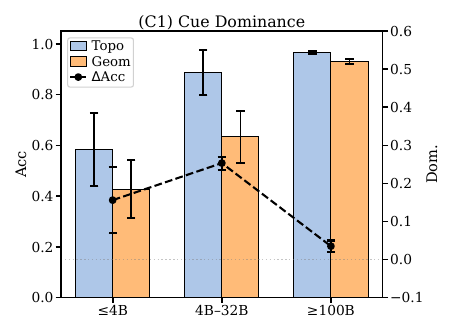}
  \caption{Cue dominance under sufficient information (C1, tier-wise).
  Bars compare Topo vs.\ Geom accuracy; dashed line shows dominance score ($\Delta Acc$).}
  \label{fig:c1_tier}
\end{figure}

As illustrated in Figure~\ref{fig:c1_tier}, models across all tiers ($\le$4B, 4B--32B, $\ge$100B) exhibit a consistent and significant preference for the topological regime. Specifically, $S_{dom}$ remains stable between $0.22$ and $0.32$ regardless of parameter scale. From a Transformer architecture perspective, this suggests that the multi-head self-attention mechanism is more adept at capturing \textit{discrete relational dependencies} (as represented in topological graphs) than performing \textit{implicit numerical mapping} of continuous coordinate spaces. The topological shield acts as a structural prior that aligns with the LLM's pre-training on structured linguistic data.

\textit{Engineering Guidance:} In embodied AI prompts, developers should prioritize graph-based topological connectivity over raw coordinate sets. Since LLMs' ability to utilize geometric information is significantly limited by their intrinsic inductive bias, explicit relational mapping is essential for reliable navigation.

\paragraph{Contextual Conflict Probing (Regime C2 Results).}
To expose the ``Achilles' heel'' of LLM spatial reasoning, we introduce semantic conflict to simulate perceptual aliasing. We define the Semantic Achilles' Heel Index (SAHI) as the relative performance loss under conflict:
\begin{equation}
    SAHI = \frac{Acc_{shield} - Acc_{conflict}}{Acc_{shield}}
\end{equation}

\begin{figure}[t]
  \centering
  \includegraphics[width=0.90\columnwidth]{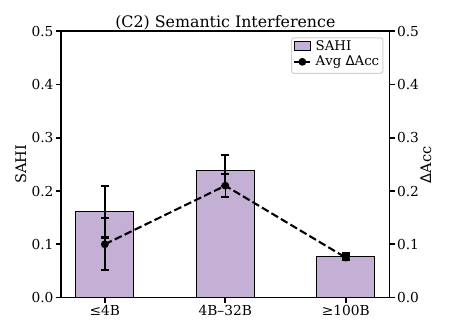}
  \caption{Semantic vulnerability under cue conflict (C2, tier-wise).
  Bars: SAHI; dashed line: mean performance drop ($\Delta Acc$).}
  \label{fig:c2_tier}
\end{figure}

As shown in Figure~\ref{fig:c2_tier}, semantic interference effectively penetrates the topological shield, especially in mid-sized models. Tier 2 models (4B--32B) exhibit the highest vulnerability, with a SAHI score of approximately $0.24$ and an average accuracy drop ($\Delta Acc_{Avg}$) exceeding $0.21$. This synchronization between relative loss (SAHI) and absolute performance collapse ($\Delta Acc_{Avg}$) indicates a ``semantic-logic decoupling''. The high-dimensional embeddings of duplicate room labels (e.g., ``Bedroom'') generate excessive attention scores that override distinct topological tokens, leading to systemic \textit{entity mis-binding} across all core reasoning tasks.

\textit{Engineering Guidance:} For real-world deployment, semantic disambiguation is mandatory. Relying solely on topological IDs is insufficient to counteract the LLM's inherent semantic bias. Scenes with duplicate labels must be augmented with unique semantic descriptors (e.g., ``North Bedroom'') to prevent the navigation system from collapsing due to label confusion.

\section{Conclusion}

LLM-based navigation systems commonly serialize explicit spatial representations into text, yet the \emph{linguistic structure} of this serialization and the \emph{contextual cues} exposed to the model are often treated as neutral engineering choices. This paper challenges this assumption by arguing that serialization itself constitutes a modeling decision that actively shapes attention, evidence weighting, and failure modes in Transformer-based planning.

To make this interface analyzable, we introduce a dual-interventional framework that disentangles (i) representation intervention over \emph{linguistic format} and \emph{degree of compression} under strict information equivalence, and (ii) context intervention that controls cue availability and injects cue conflicts to reveal planning-relevant preferences and vulnerabilities.

Across tasks and model scales, we identify a consistent pattern of linguistic inductive bias captured by three metaphors. First, the Sword: explicit linguistic structure acts as a conditional scaffold, benefiting deployment-friendly models but exhibiting scale-dependent reversals and overhead under high compression or spatial complexity. Second, the Shield: topological connectivity provides the most stable backbone for navigation planning, dominating geometry-based cues even when both are sufficient. Third, the Achilles' Heel: semantic ambiguity can override otherwise-correct structural evidence, penetrating topological robustness and systematically derailing global route construction.

These findings translate into concrete engineering guidance for LLM-based navigation: preserve topological integrity as a first-class constraint, calibrate representational compression to model capacity and spatial load, select linguistic format in a task- and deployment-aware manner, and treat semantic correctness and disambiguation as mandatory rather than auxiliary.

Our study is limited by its focus on controlled synthetic environments designed to ensure information equivalence. An important direction for future work is to extend the proposed intervention framework to real-world benchmarks with perceptual noise and incomplete annotations, testing whether the same inductive-bias signatures persist under realistic embodied settings.

\section*{Acknowledgements}
We gratefully acknowledge financial support from the National Natural Science Foundation of China (Grant No.42130112, No.42371479); the Ministry of Industry and Information Technology of China through the National Science and Technology Major Project (Grant No.2025ZD1606804, No.2025ZD1601304); the Shanghai Natural Science Foundation (General Program, Grant No.24ZR1419800, No.23ZR1419300); the Science and Technology Commission of Shanghai Municipality (Grant No.22DZ2229004); and the Shanghai Frontiers Science Center of Molecule Intelligent Syntheses.

\bibliographystyle{named}
\bibliography{ijcai26}

@inproceedings{Deitke2020,
  title     = {RoboTHOR: An Open Simulation-to-Real Embodied {AI} Platform},
  author    = {Deitke, Matt and Han, Winson and Herrasti, {\'A}lvaro and Kembhavi, Aniruddha and Kolve, Eric and Mottaghi, Roozbeh and Salvador, Jordi and Schwenk, Dustin and VanderBilt, Eli and Wallingford, Matthew and Weihs, Luca and Yatskar, Mark and Farhadi, Ali},
  booktitle = {Proceedings of the IEEE/CVF Conference on Computer Vision and Pattern Recognition (CVPR)},
  publisher = {IEEE/CVF},
  address   = {Seattle, WA},
  year      = {2020}
}

@inproceedings{Savva2019Habitat,
  title     = {Habitat: A Platform for Embodied {AI} Research},
  author    = {Savva, Manolis and Kadian, Abhishek and Maksymets, Oleksandr and Zhao, Yili and Wijmans, Erik and Jain, Bhavana and Straub, Julian and Liu, Jia and Koltun, Vladlen and Malik, Jitendra and Parikh, Devi and Batra, Dhruv},
  booktitle = {Proceedings of the IEEE/CVF International Conference on Computer Vision (ICCV)},
  publisher = {IEEE/CVF},
  address   = {Seoul, Korea},
  year      = {2019},
  note      = {arXiv:1904.01201}
}

@article{Wei2022,
  title   = {Chain-of-Thought Prompting Elicits Reasoning in Large Language Models},
  author  = {Wei, Jason and Wang, Xuezhi and Schuurmans, Dale and Bosma, Maarten and Ichter, Brian and Xia, Fei and Chi, Ed and Le, Quoc and Zhou, Denny},
  journal = {arXiv preprint arXiv:2201.11903},
  year    = {2022}
}

@article{Yao2022ReAct,
  title   = {ReAct: Synergizing Reasoning and Acting in Language Models},
  author  = {Yao, Shunyu and Zhao, Jeffrey and Yu, Dian and Du, Nan and Shafran, Izhak and Narasimhan, Karthik and Cao, Yuan},
  journal = {arXiv preprint arXiv:2210.03629},
  year    = {2022}
}

@article{Wang2023Voyager,
  title   = {Voyager: An Open-Ended Embodied Agent with Large Language Models},
  author  = {Wang, Guanzhi and Xie, Yuqi and Jiang, Yunfan and Mandlekar, Ajay and Xiao, Chaowei and Zhu, Yuke and Fan, Linxi and Anandkumar, Anima},
  journal = {arXiv preprint arXiv:2305.16291},
  year    = {2023}
}

@inproceedings{Shah2023LMNav,
  title     = {LM-Nav: Robotic Navigation with Large Pre-Trained Models of Language, Vision, and Action},
  author    = {Shah, Dhruv and Osinski, Blazej and Ichter, Brian and Levine, Sergey},
  booktitle = {Proceedings of the Conference on Robot Learning (CoRL)},
  publisher = {PMLR},
  address   = {Atlanta, GA},
  year      = {2023},
  note      = {Also available as arXiv:2207.04429}
}

@article{Huang2023VLMaps,
  title   = {Visual Language Maps for Robot Navigation},
  author  = {Huang, Chenguang and Mees, Oier and Zeng, Andy and Burgard, Wolfram},
  journal = {arXiv preprint arXiv:2210.05714},
  year    = {2022},
  note    = {Also appeared in ICRA 2023}
}

@article{Rana2023SayPlan,
  title   = {SayPlan: Grounding Large Language Models using 3D Scene Graphs for Scalable Robot Task Planning},
  author  = {Rana, Kanishka and Haviland, Jack and others},
  journal = {arXiv preprint arXiv:2307.06135},
  year    = {2023},
  note    = {Also appeared in CoRL 2023}
}

@article{Chen2024MapGPT,
  title   = {MapGPT: Map-Guided Prompting with Adaptive Path Planning for Vision-and-Language Navigation},
  author  = {Chen, Jiaqi and Lin, Bingqian and Xu, Ran and Chai, Zhenhua and Liang, Xiaodan and Wong, Kwan-Yee K.},
  journal = {arXiv preprint arXiv:2401.07314},
  year    = {2024},
  note    = {Also appeared in ACL 2024}
}

@article{Zhou2023NavGPT,
  title   = {NavGPT: Explicit Reasoning in Vision-and-Language Navigation with Large Language Models},
  author  = {Zhou, Gengze and Hong, Yicong and Wu, Qi},
  journal = {arXiv preprint arXiv:2305.16986},
  year    = {2023}
}

@article{Zhang2025MapNav,
  title   = {MapNav: A Novel Memory Representation via Annotated Semantic Maps for VLM-based Vision-and-Language Navigation},
  author  = {Zhang, Lingfeng and Hao, Xiaoshuai and Xu, Qinwen and Zhang, Qiang and Zhang, Xinyao and Wang, Pengwei and Zhang, Jing and Wang, Zhongyuan and Zhang, Shanghang and Xu, Renjing},
  journal = {arXiv preprint arXiv:2502.13451},
  year    = {2025}
}

@article{Vaswani2017,
  title   = {Attention Is All You Need},
  author  = {Vaswani, Ashish and Shazeer, Noam and Parmar, Niki and Uszkoreit, Jakob and Jones, Llion and Gomez, Aidan N. and Kaiser, Lukasz and Polosukhin, Illia},
  journal = {arXiv preprint arXiv:1706.03762},
  year    = {2017}
}

@article{Liu2023LostInMiddle,
  title   = {Lost in the Middle: How Language Models Use Long Contexts},
  author  = {Liu, Nelson F. and others},
  journal = {arXiv preprint arXiv:2307.03172},
  year    = {2023},
  note    = {Also appeared in TACL 2024}
}

@article{Fatemi2023Talk2Graph,
  title   = {Talk like a Graph: Encoding Graphs for Large Language Models},
  author  = {Fatemi, Bahare and others},
  journal = {arXiv preprint arXiv:2310.04560},
  year    = {2023}
}

@article{Tolman1948,
  title   = {Cognitive Maps in Rats and Men},
  author  = {Tolman, Edward C.},
  journal = {Psychological Review},
  volume  = {55},
  number  = {4},
  pages   = {189--208},
  year    = {1948}
}

@incollection{Montello1998Framework,
  title     = {A New Framework for Understanding the Acquisition of Spatial Knowledge in Large-Scale Environments},
  author    = {Montello, Daniel R.},
  booktitle = {Spatial and Temporal Reasoning in Geographic Information Systems},
  editor    = {Egenhofer, Max J. and Golledge, Reginald G.},
  publisher = {Oxford University Press},
  pages     = {143--154},
  year      = {1998}
}

@incollection{Montello2001SpatialCognition,
  title     = {Spatial Cognition},
  author    = {Montello, Daniel R.},
  booktitle = {International Encyclopedia of the Social \& Behavioral Sciences},
  editor    = {Smelser, Neil J. and Baltes, Paul B.},
  publisher = {Elsevier},
  pages     = {14771--14775},
  year      = {2001},
  note      = {DOI:10.1016/B0-08-043076-7/02492-X}
}

@incollection{MontelloSas2006Wayfinding,
  title     = {Human Factors of Wayfinding in Navigation},
  author    = {Montello, Daniel R. and Sas, Corina},
  booktitle = {International Encyclopedia of Ergonomics and Human Factors},
  publisher = {CRC Press/Taylor \& Francis},
  year      = {2006},
  pages     = {2003--2008}
}

@article{HirtleJonides1985,
  title   = {Evidence of Hierarchies in Cognitive Maps},
  author  = {Hirtle, Stephen C. and Jonides, John},
  journal = {Memory \& Cognition},
  volume  = {13},
  number  = {3},
  pages   = {208--217},
  year    = {1985}
}

@article{Kojima2022,
  title   = {Large Language Models are Zero-Shot Reasoners},
  author  = {Kojima, Takeshi and Gu, Shixiang Shane and Reid, Machel and Matsuo, Yutaka and Iwasawa, Yusuke},
  journal = {arXiv preprint arXiv:2205.11916},
  year    = {2022}
}

@article{Kaplan2020Scaling,
  title   = {Scaling Laws for Neural Language Models},
  author  = {Kaplan, Jared and McCandlish, Sam and Henighan, Tom and Brown, Tom B. and Chess, Benjamin and Child, Rewon and Gray, Scott and Radford, Alec and Wu, Jeffrey and Amodei, Dario},
  journal = {arXiv preprint arXiv:2001.08361},
  year    = {2020}
}

@book{Thrun2005,
  title     = {Probabilistic Robotics},
  author    = {Thrun, Sebastian and Burgard, Wolfram and Fox, Dieter},
  publisher = {MIT Press},
  address   = {Cambridge, Massachusetts},
  year      = {2005}
}

@article{Kuipers2000,
  title   = {The Spatial Semantic Hierarchy},
  author  = {Kuipers, Benjamin},
  journal = {Artificial Intelligence},
  volume  = {119},
  number  = {1--2},
  pages   = {191--233},
  year    = {2000}
}

@book{Latombe1991,
  title     = {Robot Motion Planning},
  author    = {Latombe, Jean-Claude},
  publisher = {Kluwer Academic Publishers},
  address   = {Norwell, MA},
  year      = {1991}
}

@article{Liang2022CodeAsPolicies,
  title   = {Code as Policies: Language Model Programs for Embodied Control},
  author  = {Liang, Jacky and others},
  journal = {arXiv preprint arXiv:2209.07753},
  year    = {2022}
}

@article{Wang2023NLGraph,
  title   = {Can Language Models Solve Graph Problems in Natural Language?},
  author  = {Wang, Heng and Feng, Shangbin and He, Tianxing and Tan, Zhaoxuan and Han, Xiaochuang and Tsvetkov, Yulia},
  journal = {arXiv preprint arXiv:2305.10037},
  year    = {2023}
}

@inproceedings{Valmeekam2023,
  title     = {On the Planning Abilities of Large Language Models (A Critical Investigation)},
  author    = {Valmeekam, Karthik and others},
  booktitle = {Advances in Neural Information Processing Systems (NeurIPS)},
  publisher = {Curran Associates},
  address   = {New Orleans, LA},
  year      = {2023},
  note      = {arXiv:2310.12397}
}

@article{Zhang2024GraphLLM,
  title   = {Can {LLM} Graph Reasoning Generalize beyond Pattern Memorization?},
  author  = {Zhang, Y. and others},
  journal = {arXiv preprint arXiv:2406.15992},
  year    = {2024}
}

@article{Gurnee2023SpaceTime,
  title   = {Language Models Represent Space and Time},
  author  = {Gurnee, Wes and Tegmark, Max},
  journal = {arXiv preprint arXiv:2310.02207},
  year    = {2023}
}

@article{Bubeck2023,
  title   = {Sparks of Artificial General Intelligence: Early Experiments with {GPT}-4},
  author  = {Bubeck, S{\'e}bastien and Chandrasekaran, Varun and Eldan, Ronen and Gehrke, Johannes and Horvitz, Eric and Kamar, Ece and Lee, Peter and Lee, Yin Tat and Li, Yuanzhi and Lundberg, Scott and Nori, Harsha and Palangi, Hamid and Ribeiro, Marco Tulio and Zhang, Yi},
  journal = {arXiv preprint arXiv:2303.12712},
  year    = {2023}
}

@article{Cadena2016past,
  title   = {Past, Present, and Future of Simultaneous Localization and Mapping: Toward the Robust-Perception Age},
  author  = {Cadena, Cesar and Carlone, Luca and Carrillo, Henry and Latif, Yasir and Scaramuzza, Davide and Neira, Jos{\'e} and Reid, Ian and Leonard, John J.},
  journal = {IEEE Transactions on Robotics},
  volume  = {32},
  number  = {6},
  pages   = {1309--1332},
  year    = {2016}
}

@article{touvron2023llama,
  title   = {{LLaMA}: Open and Efficient Foundation Language Models},
  author  = {Touvron, Hugo and Lavril, Thibaut and Izacard, Gautier and others},
  journal = {arXiv preprint arXiv:2302.13971},
  year    = {2023}
}

@article{grattafiori2024llama3,
  title   = {The {Llama} 3 Herd of Models},
  author  = {Grattafiori, Aaron and others},
  journal = {arXiv preprint arXiv:2407.21783},
  year    = {2024}
}

@article{bai2023qwen,
  title   = {Qwen Technical Report},
  author  = {Bai, Jinze and Bai, Shuai and Chu, Yunfei and others},
  journal = {arXiv preprint arXiv:2309.16609},
  year    = {2023}
}

@article{qwen2024qwen2,
  title   = {Qwen2 Technical Report},
  author  = {Yang, An and others},
  journal = {arXiv preprint arXiv:2407.10671},
  year    = {2024}
}

@article{Ahn2022,
  title   = {Do As I Can, Not As I Say: Grounding Language in Robotic Affordances},
  author  = {Ahn, Michael and Brohan, Anthony and Brown, Noah and Chebotar, Yevgen and Cortes, Omar and David, Byron and others},
  journal = {arXiv preprint arXiv:2204.01691},
  year    = {2022}
}

@article{Huang2022InnerMonologue,
  title   = {Inner Monologue: Embodied Reasoning through Planning with Language Models},
  author  = {Huang, Wenlong and Abbeel, Pieter and Pathak, Deepak and others},
  journal = {arXiv preprint arXiv:2207.05608},
  year    = {2022}
}

@article{Zhang2017NeuralSLAM,
  title   = {Neural {SLAM}: Learning to Explore with External Memory},
  author  = {Zhang, Jingwei and Tai, Lei and Boedecker, Joschka and Burgard, Wolfram and Liu, Ming},
  journal = {arXiv preprint arXiv:1706.09520},
  year    = {2017}
}

@article{Parisotto2017NeuralMap,
  title   = {Neural Map: Structured Memory for Deep Reinforcement Learning},
  author  = {Parisotto, Emilio and Salakhutdinov, Ruslan},
  journal = {arXiv preprint arXiv:1702.08360},
  year    = {2017}
}

\end{document}